\documentclass[10pt,twocolumn,letterpaper]{article}

\usepackage{cvpr}
\usepackage{times}
\usepackage{epsfig}
\usepackage{graphicx}
\usepackage{amsmath}
\usepackage{amssymb}
\usepackage{color}
\usepackage{url}
\usepackage{subfigure}
\usepackage{color}
\usepackage{multirow}
\usepackage[pagebackref=true,breaklinks=true,letterpaper=true,hyperfootnotes=true,colorlinks,bookmarks=false]{hyperref}

\usepackage{booktabs}
\usepackage{pifont}
\usepackage{makecell}

\usepackage{authblk}

\cvprfinalcopy 

\def\cvprPaperID{2075} 

\ifcvprfinal\fi
\begin{document}

\title{3DV: 3D Dynamic Voxel for Action Recognition in Depth Video}

\author{\vspace{-4 mm}  Yancheng~Wang$^1$, Yang~Xiao$^1$$^\dag$, Fu~Xiong$^2$, Wenxiang~Jiang$^1$, Zhiguo~Cao$^1$, Joey Tianyi~Zhou$^3$, and Junsong Yuan$^4$ \\
 \vspace{-4 mm} $^1$  National Key Laboratory of Science and Technology on Multispectral Information Processing, School of Artificial Intelligence and Automation, Huazhong University of Science and Technology, Wuhan 430074, China~~~~~~~$^2$Megvii Research Nanjing, Megvii Technology, China\\
$^3$ IHPC, A*STAR, Singapore~~~~~~~$^4$CSE Department, State University of New York at Buffalo\\
\tt\small{yancheng$\_$wang, Yang$\_$Xiao}@hust.edu.cn, xiongfu@megvii.com, {wenx$\_$jiang, zgcao}@hust.edu.cn,\\ \tt\small zhouty@ihpc.a-star.edu.sg, jsyuan@buffalo.edu
}

\maketitle

\let\thefootnote\relax\footnotetext{\dag Yang Xiao is corresponding author (Yang$\_$Xiao@hust.edu.cn).}

\begin{abstract}

To facilitate depth-based 3D action recognition, 3D dynamic voxel (3DV) is proposed as a novel 3D motion representation. With 3D space voxelization, the key idea of 3DV is to encode 3D motion information within depth video into a regular voxel set (i.e., 3DV) compactly, via temporal rank pooling. Each available 3DV voxel intrinsically involves 3D spatial and motion feature jointly. 3DV is then abstracted as a point set and input into PointNet++ for 3D action recognition, in the end-to-end learning way. The intuition for transferring 3DV into the point set form is that, PointNet++ is lightweight and effective for deep feature learning towards point set. Since 3DV may lose appearance clue, a multi-stream 3D action recognition manner is also proposed to learn motion and appearance feature jointly. To extract richer temporal order information of actions, we also divide the depth video into temporal splits and encode this procedure in 3DV integrally. The extensive experiments on 4 well-established benchmark datasets demonstrate the superiority of our proposition. Impressively, we acquire the accuracy of $82.4\%$ and $93.5\%$ on NTU RGB+D 120~\cite{liu2019ntu} with the cross-subject and cross-setup test setting respectively. 3DV's code is available at \url{https://github.com/3huo/3DV-Action}.

\end{abstract}

\begin{figure}
\centering
\includegraphics[width=0.35\textwidth]{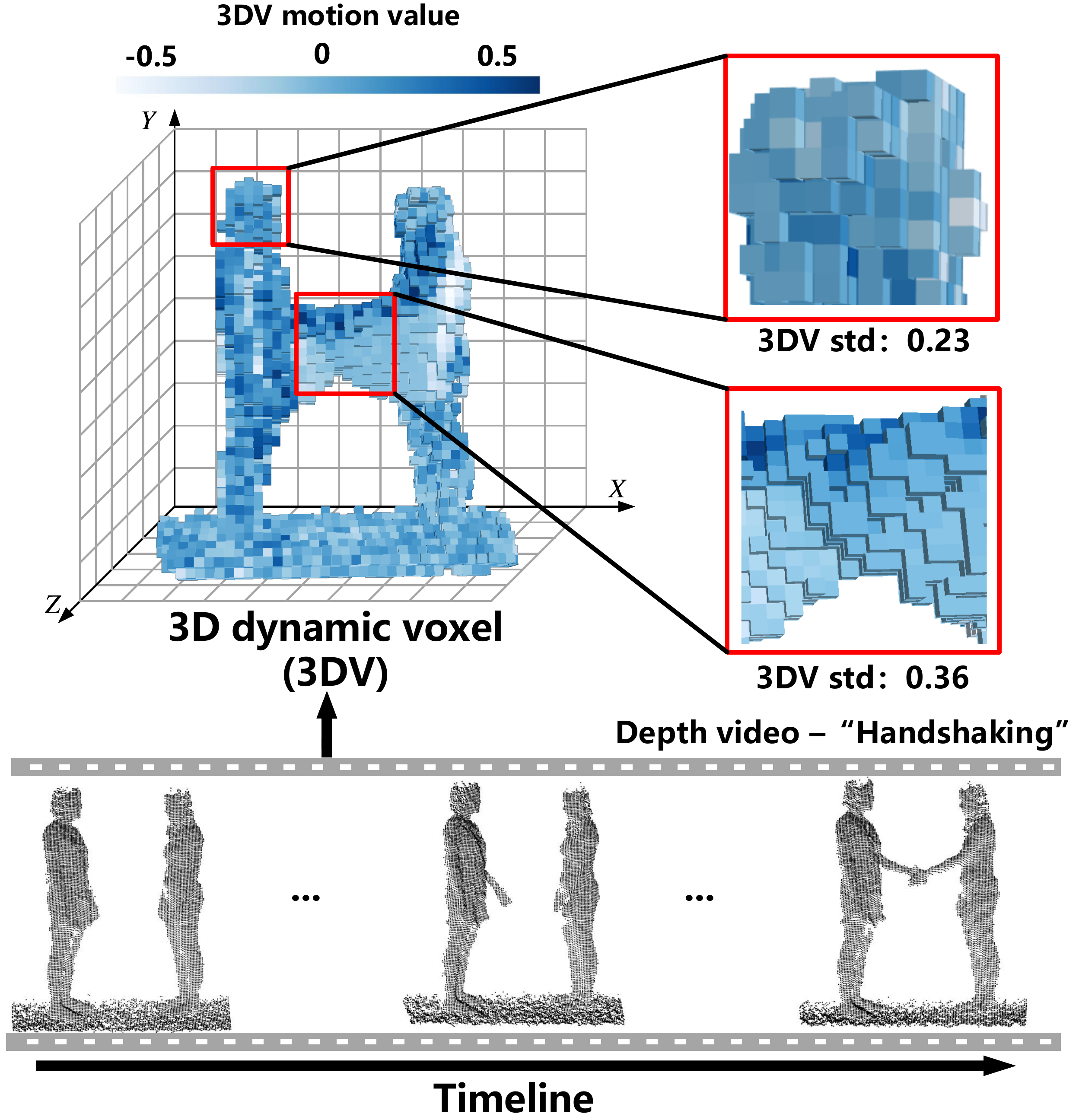}
\caption{A live ``Handshaking" 3DV example from NTU RGB+D 60 dataset~\cite{shahroudy2016ntu}. 3DV motion value reveals the temporal order of 3D motion component. The later motion component is of higher value, and vice verse. And, the local region of richer motion information holds higher standard deviation on 3DV motion value.}
\label{fig:3dv}
\vspace{-0.2cm}
\end{figure}

\section{Introduction} \label{introduction}

During the past decade, due to the emergence of low-cost depth camera (e.g., Microsoft Kinect~\cite{zhang2012microsoft}) 3D action recognition becomes an active research topic, with the wide-range application scenarios of video surveillance, human-machine interaction, etc~\cite{wang2018rgb,xiao2019action}. The state-of-the-art 3D action recognition approaches can be generally categorized into the depth-based~\cite{rahmani20163d,liu2017global,liu2017skeleton,si2019attention,li2019actional,zhang2019view,shi2019two,shi2019skeleton} and skeleton-based groups~\cite{ren2013robust,yang2014super,ohn2013joint,li2018unsupervised,Wang2018Depth,xiao2019action}. Since accurate and robust 3D human pose estimation is still challenging~\cite{xiong2019a2j,moon2018v2v}, we focus on depth-based avenue in this work.

Since human conducts actions in 3D space, capturing 3D motion pattern effectively and efficiently is crucial for depth-based 3D action recognition. An intuitive way is to calculate dense scene flow~\cite{basha2013multi}. However this can be time consuming~\cite{basha2013multi}, which may not be preferred by the practical applications. Recently, dynamic image~\cite{bilen2016dynamic,bilen2017action} able to represent the motion information within RGB video compactly has been introduced to depth domain for 3D action characterization~\cite{Wang2018Depth,xiao2019action}. It can compress RGB video into a single image, while still maintaining the motion characteristics well via temporal rank pooling~\cite{fernando2015modeling,fernando2016rank}. Thus dynamic image can fit deep CNN model~\cite{he2016deep} well for action categorization, which is leveraged by CNN's strong pattern representation capacity. Nevertheless we argue that the ways of applying dynamic image to 3D field in~\cite{Wang2018Depth,xiao2019action} have not fully exploited 3D descriptive clue within depth video, although normal vector~\cite{Wang2018Depth} or multi-view projection~\cite{xiao2019action} is applied. The insight is that, both methods in~\cite{Wang2018Depth,xiao2019action} finally encode 3D motion information onto the 2D image plane to fit CNN. Thus, they cannot well answer the question \emph{``Where does the certain 3D motion pattern within human action appear in 3D space? " } crucial for effective 3D action characterization due to the fact that human actions actually consists of both motion patterns and compact spatial structure~\cite{rao2002view}.

To address the concern above, we propose 3D dynamic voxel (3DV) as a novel 3D motion representation for 3D action representation. To extract 3DV, 3D  space voxelization is first executed. Each depth frame will be transformed into a regular voxel set. And the appearance content within it can be encoded by observing whether the yielded voxels have been occupied or not~\cite{wang2012robust}, in a binary way. Then, temporal rank pooling~\cite{fernando2015modeling,fernando2016rank} is executed towards all the binary voxel sets to compress them into one single voxel set termed 3DV particularly. Thus, \emph{3D motion and spatial characteristics of 3D action can be encoded into 3DV jointly.} To reveal this, a live ``Handshaking" 3DV example is provided in Fig.~\ref{fig:3dv}. As shown, each available 3DV voxel possesses a motion value able to reflect the temporal order of its corresponding 3D motion component. Specifically, the later motion component is of higher value, and vice verse. Meanwhile, the local region of richer 3D motion information possesses higher standard deviation on 3DV motion value (e.g., hand region vs. head region). Meanwhile, 3DV voxel's location reveals the 3D position of its 3D motion component. Thus, 3DV's spatial-motion representative ability can essentially leverage 3D action characterization. To involve richer temporal order information, we further divide depth video into finer temporal splits. This is encoded in 3DV integrally by fusing the motion values from all the temporal splits.

With 3DV, the upcoming question is how to choose the adaptive deep learning model to conduct 3D action recognition particularly. Towards voxel set, 3D CNN~\cite{maturana2015voxnet,ge20173d,moon2018v2v} is often used for 3D visual pattern understanding, and also applicable to 3DV. However, it is difficult to train due to the large number of convolutional parameters. Inspired by the recent success of the lightweighted deep learning models on point set (e.g., PointNet++~\cite{qi2017pointnet++}), we propose to transfer 3DV into the point set form as the input of PointNet++ to conduct 3D action recognition in end-to-end learning manner. That is, each 3DV voxel will be abstract as a point characterized by its 3D location index and motion value. Our intuition is to alleviate the training difficulty and burden.


Although 3DV can reveal 3D motion information, it still may lose appearance details as in Fig.~\ref{fig:3dv}. Since appearance also plays vital role for action recognition~\cite{oreifej2013hon4d,simonyan2014two}, only using 3DV may weaken performance. To alleviate this, a multi-stream deep learning model using PointNet++ is also proposed to learn 3D motion and appearance feature jointly. In particular, it consists of one motion stream and multiple appearance streams. The input of motion stream is 3DV. And, the inputs of appearance streams are the depth frames sampled from the different temporal splits. They will also be transformed into the point set form to fit PointNet++.

{The experiments on 2 large-scale 3D action recognition datasets (i.e., NTU RGB+D 120~\cite{liu2019ntu} and 60~\cite{shahroudy2016ntu}), and 2 small-scale ones (i.e., N-UCLA~\cite{wang2014cross} and UWA3DII~\cite{rahmani2016histogram}) verify 3DV's superiority over the state-of-the-art manners.}

The main contributions of this paper include:

$\bullet$ 3DV: a novel and compact 3D motion representative manner for 3D action characterization;

$\bullet$ PointNet++ is applied to 3DV for 3D action recognition in end-to-end learning way, from point set perspective;

$\bullet$ A multi-stream deep learning model is proposed to learn 3D motion and appearance feature jointly.

\begin{figure}[t]
\centering
\subfigure[Human-object interaction] {\label{sub_hand} \includegraphics[height=2.4cm,width=3.3cm]{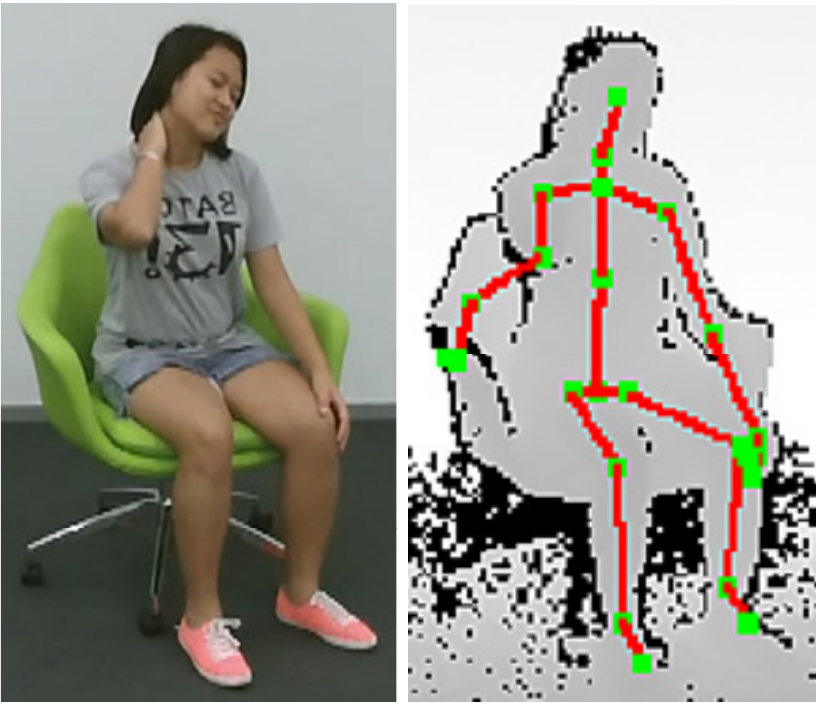}}
\subfigure[Self-occlusion] {\label{sub_human} \includegraphics[height=2.4cm]{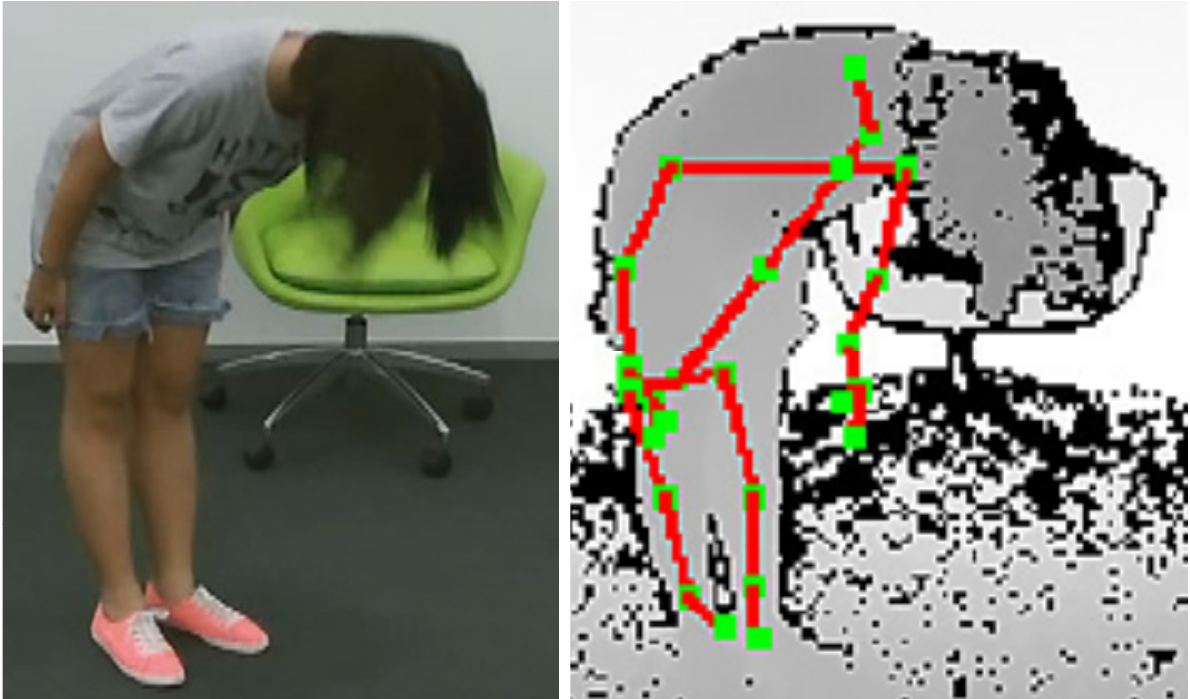}}

\caption{3D skeleton extraction failure cases in NTU RGB+D 60 dataset~\cite{liu2019ntu}, due to human-object interaction and self-occlusion. The depth frame and its RGB counterpart are shown jointly.}
\label{fig:failure_case}
\vspace{-0.2cm}
\end{figure}
\section{Related Works} \label{related-work}

\textbf{3D action recognition}. The existing 3D action recognition approaches generally falls into the depth-based~\cite{oreifej2013hon4d,oreifej2013hon4d,yang2014super,ohn2013joint,li2018unsupervised,Wang2018Depth,xiao2019action} and skeleton-based groups~\cite{liu2016spatio,liu2017global,liu2017skeleton,si2019attention,li2019actional,zhang2019view,shi2019two,shi2019skeleton}. Recently the skeleton-based approaches with RNN~\cite{liu2016spatio} and GCN~\cite{shi2019two} has drawn more attention, since using 3D skeleton can help to resist the impact of variations on scene, human attribute, imaging viewpoint, etc. However, one critical issue should not be ignored. That is, accurate and robust 3D human pose estimation is still not trivial~\cite{xiong2019a2j,moon2018v2v}. To reveal this, we have checked the 3D skeletons within NTU RGB+D 60~\cite{liu2019ntu} carefully. Actually, even under the constrained condition 3D skeleton extraction still may fail to work as in Fig.~\ref{fig:failure_case}. Thus, currently for the practical applications depth-based manner seems more preferred and is what we concern.

Most of the paid efforts focus on proposing 3D action representation manner to capture 3D spatial-temporal appearance or motion pattern. At the early stage, the hand-crafted descriptions of bag of 3D points~\cite{li2008expandable}, depth motion map (DMM)~\cite{yang2012recognizing}, Histogram of Oriented 4D Normals (HON4D)~\cite{oreifej2013hon4d}, Super Normal Vector (SNV)~\cite{lu2014range} and binary range sample feature~\cite{lu2014range} are proposed from the different research perspectives. Recently CNN~\cite{he2016deep,zhang2019nonlinear} has been introduced to this field~\cite{wang2015convnets,wang2015action,Wang2018Depth,xiao2019action}, and enhanced performance remarkably. Under this paradigm, the depth video will be compressed into one image using DMM~\cite{yang2012recognizing} or dynamic image~\cite{bilen2016dynamic,bilen2017action} to fit CNN. To better exploit 3D descriptive clue, normal vector or multi-view projection is applied additionally. However, they generally suffer from 2 main defects. First, as aforementioned DMM or dynamic image cannot fully reveal 3D motion characteristics well. Secondly, they tend to ignore appearance information.

\textbf{Temporal rank pooling}. To represent action, temporal rank pooling~\cite{fernando2015modeling,fernando2016rank} is proposed to capture the frame-level evolution characteristics within video. Its key idea is to train a linear ranking machine towards the frames to arrange them in chronological order. Then, the parameters of the ranking machine can be used as the action representation. By applying temporal rank pooling to the raw frame pixels, dynamic image~\cite{bilen2016dynamic,bilen2017action} is proposed with strong motion representative ability and adaptive to CNN. As aforementioned, temporal rank pooling has recently been applied to 3D action recognition~\cite{Wang2018Depth,xiao2019action}. However, how to use it to fully reveal 3D motion property still has not been deeply studied.

\textbf{Deep learning on point set}. Due to the irregularity of 3D point set, typical convolutional architectures (e.g., CNN~\cite{he2016deep}) cannot handle it well. To address this, deep learning on point set draws the increasing attention. Among the paid efforts, PointNet++~\cite{qi2017pointnet++} is the representative one. It contributes to ensure the permutation invariance of point sets, and capture 3D local geometric clue. However, it has not been applied to 3D action recognition yet.

Accordingly, 3DV is proposed to characterize 3D motion compactly, via temporal rank pooling. The adaptive multi-stream deep learning model using PointNet++ is also proposed to learn 3D motion and appearance feature jointly.


\section{3DV: A Novel Voxel Set based Compact 3D Motion Representative Manner} \label{sec:3dv}
\begin{figure}
\centering
\includegraphics[width=0.42\textwidth]{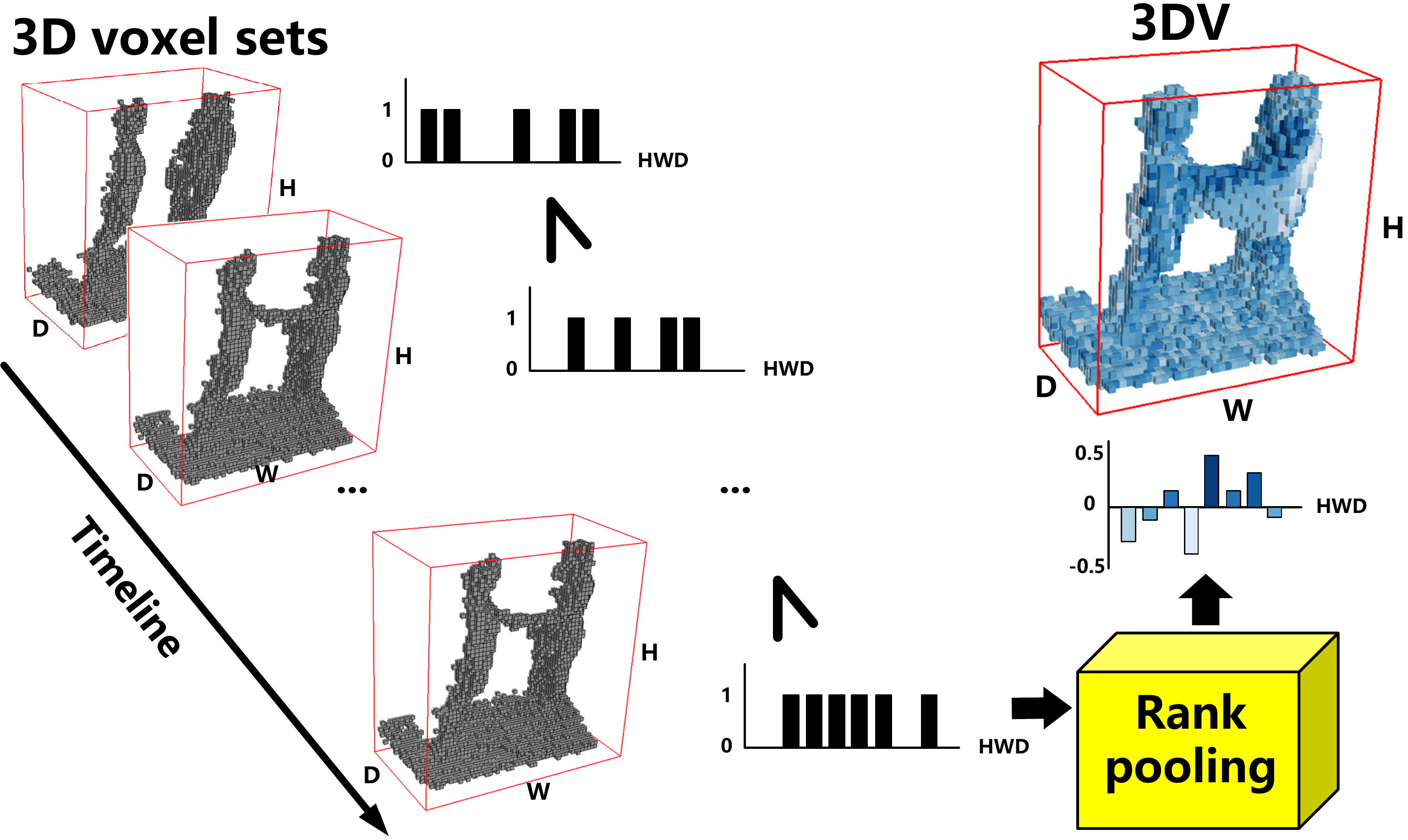}
\caption{The main idea for 3DV extraction via temporal rank pooling, towards the 3D voxel sets transformed from depth frames.}
\label{fig:3dv_construction}
\vspace{-0.2cm}
\end{figure}

Our research motivation on 3DV is to seek a compact 3D motion representative manner to characterize 3D action. Accordingly, deep feature learning can be easily conducted on it. The proposition of 3DV can be regarded as the essential effort for extending temporal rank pooling~\cite{fernando2015modeling,fernando2016rank} originally for 2D video to 3D domain, to capture 3D motion pattern and spatial clue jointly. The main idea for 3DV extraction is in Fig.~\ref{fig:3dv_construction}. The depth frames will be first map into point clouds to better reveal 3D characteristics. Then, 3D voxelization is executed to further transform the disordered point clouds into the regular voxel sets. Consequently, 3D action appearance clue within the certain depth frame can be described by judging whether the voxels have been occupied or not. Then temporal rank pooling is executed to the yielded binary voxel sets to compress them into one voxel set (i.e., 3DV), to reveal the 3D appearance evolution within actions compactly. The resulting ranking machine parameters can actually characterize 3D motion pattern of the corresponding 3DV voxels. In particular, each 3DV voxel possesses a motion value (i.e., ranking machine parameter). And, its 3D position can encode the spatial property of the corresponding 3D motion pattern. Action proposal will also be conducted to resist background.

\begin{figure}[t]
\centering  
\subfigure[Point cloud] {\label{fig:pointcloud}
\includegraphics[height=2.8cm]{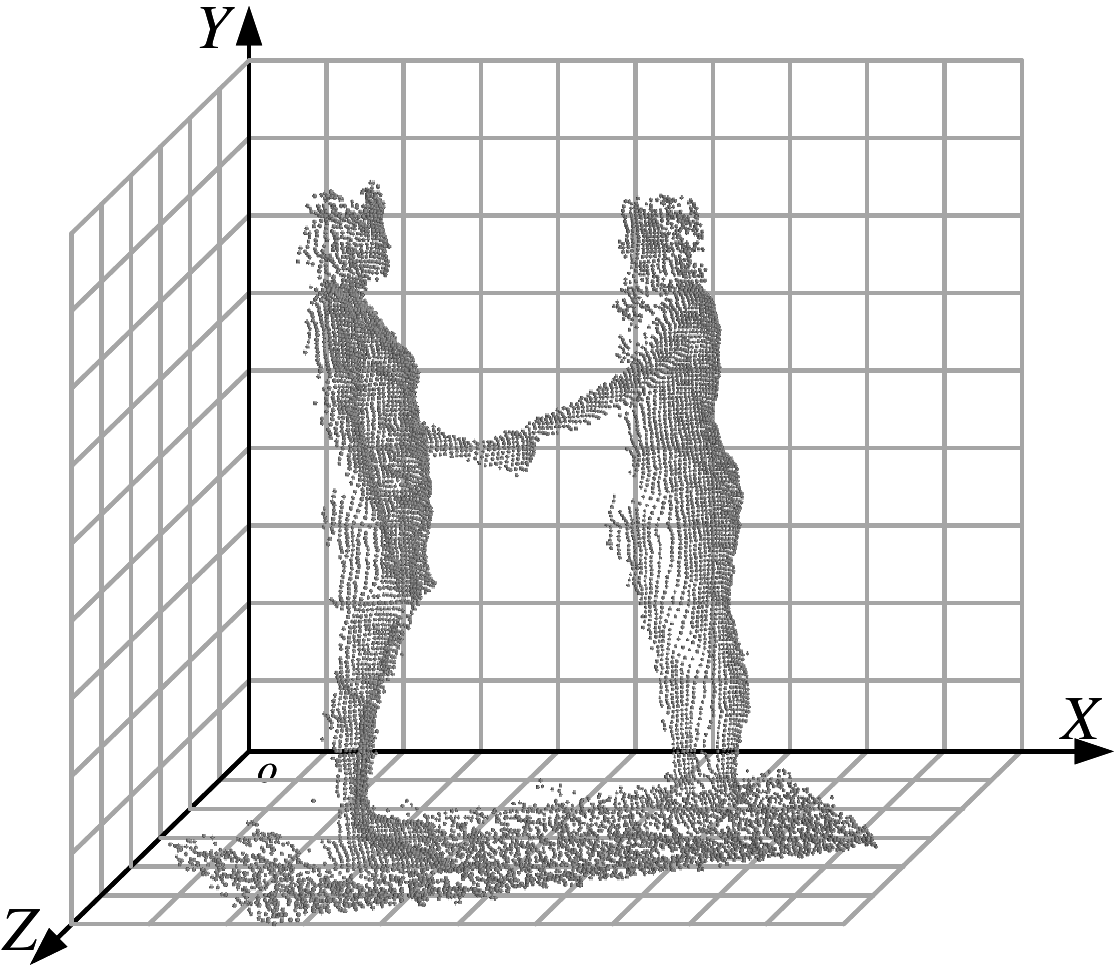}}
\subfigure[3D voxel set] {\label{fig:3dvoxel}
\includegraphics[height=2.8cm]{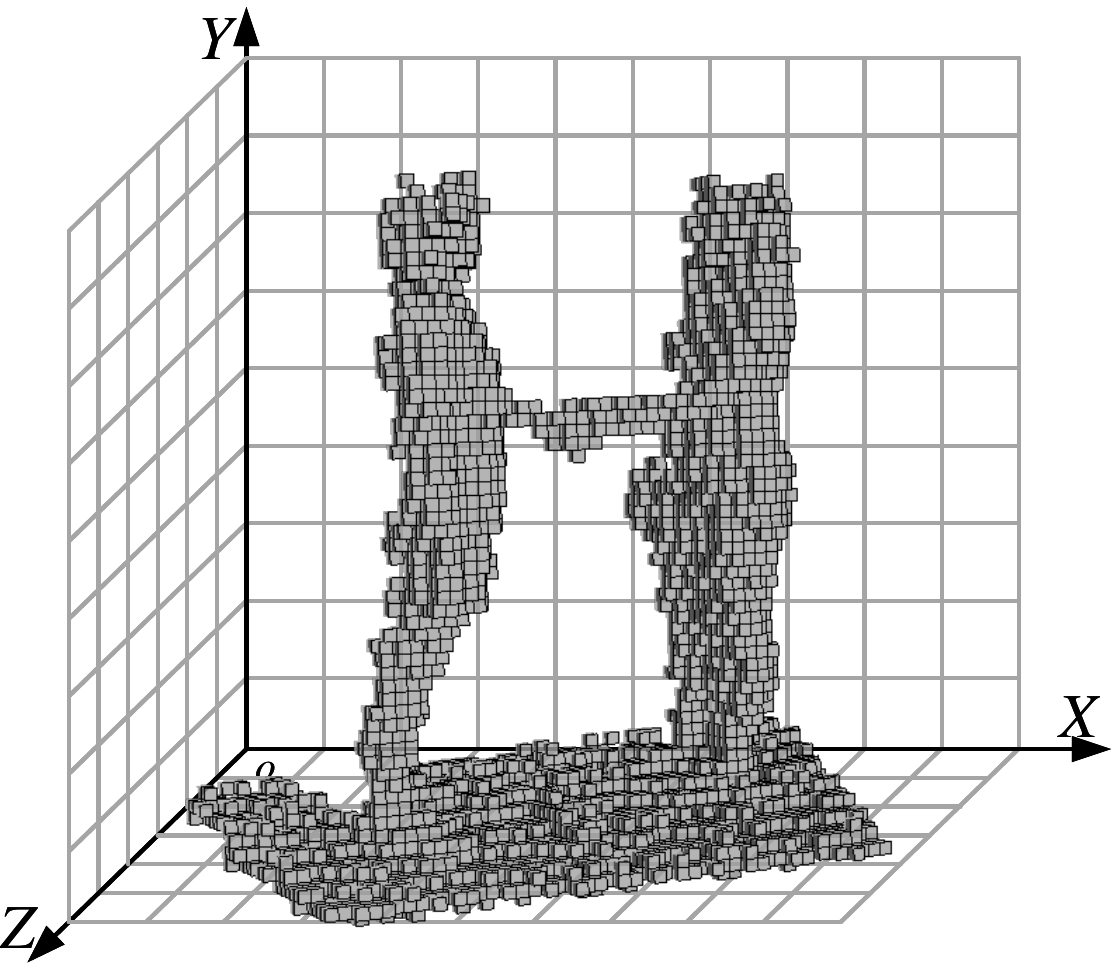}}
\caption{The point cloud and its corresponding 3D voxel set sampled from ``Handshaking".}
\label{fig:pointcloud_voxel}
\vspace{-0.2cm}
\end{figure}

\subsection{Voxel-based 3D appearance representation}

Projecting 3D data to 2D depth frame actually distorts the real 3D shape~\cite{moon2018v2v}. To better represent 3D appearance clue, we map the depth frame into point cloud. Nevertheless, one critical problem emerges. That is, temporal rank pooling cannot be applied to the yielded point clouds directly, due to their disordered property~\cite{qi2017pointnet++} as in Fig.~\ref{fig:pointcloud}. To address this, we propose to execute 3D voxelization towards the point clouds. Then the 3D appearance information can be described by observing whether the voxels have been occupied or not, disregarding the involved point number as
\begin{equation}
V_t \left(x,y,z\right)=\left\{\begin{matrix}
 1,& \text{if}\ V_t \left(x,y,z\right) \text{is occupied}\\
 0,& \text{otherwise}
\end{matrix}\right. \,,
\label{eq:binary_voxel}
\end{equation}
where $V_t \left(x,y,z\right)$ indicates one certain voxel at the $t$-th frame; $\left(x,y,z\right)$ is the regular 3D position index. This actually holds 2 main profits. First, the yielded binary 3D voxel sets are regular as in Fig.~\ref{fig:3dvoxel}. Thus, temporal rank pooling can be applied to them for 3DV extraction. Meanwhile the binary voxel-wise representation manner is of higher tolerance towards the intrinsic sparsity and density variability problem~\cite{qi2017pointnet++} within point clouds, which essentially helps to leverage generalization power.

\begin{figure}[t]
\centering  
\subfigure[Bow] {\label{fig:di_examplep(d)}
\includegraphics[height=2.4cm]{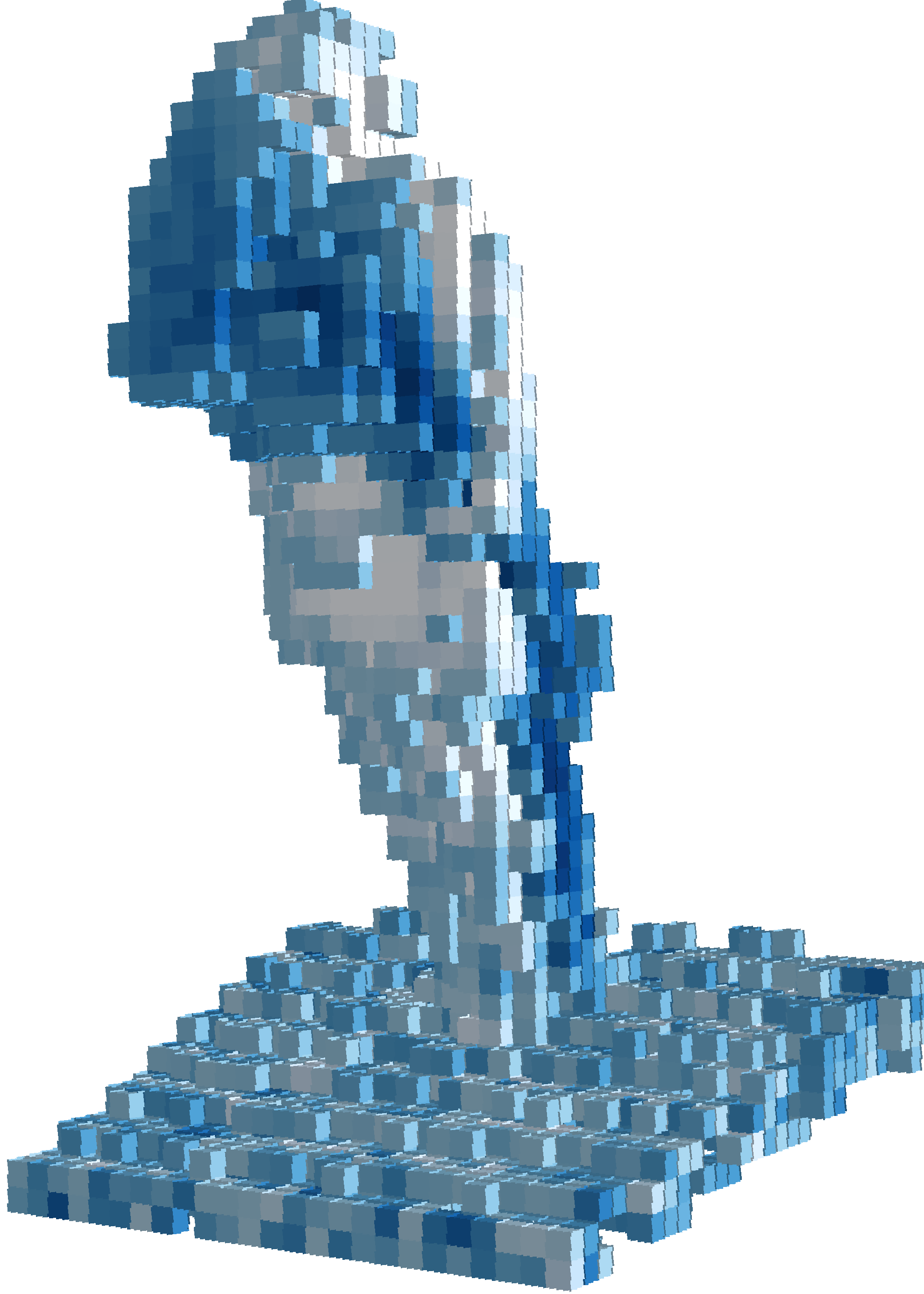}}
\subfigure[Sit down] {\label{fig:di_example(b)}
\includegraphics[height=2.4cm]{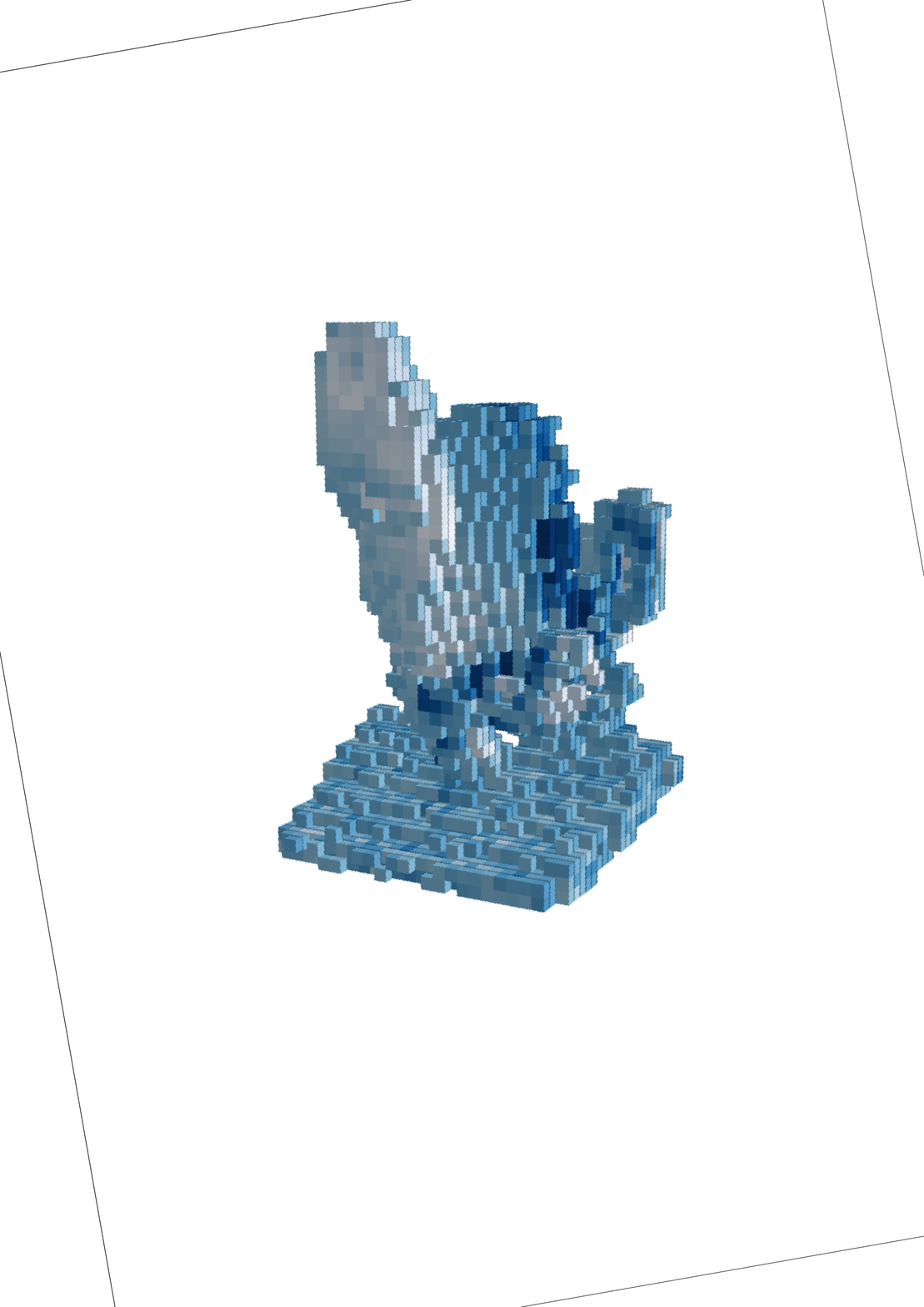}}
\subfigure[Hugging] {\label{fig:di_example(a)}
\includegraphics[height=2.4cm]{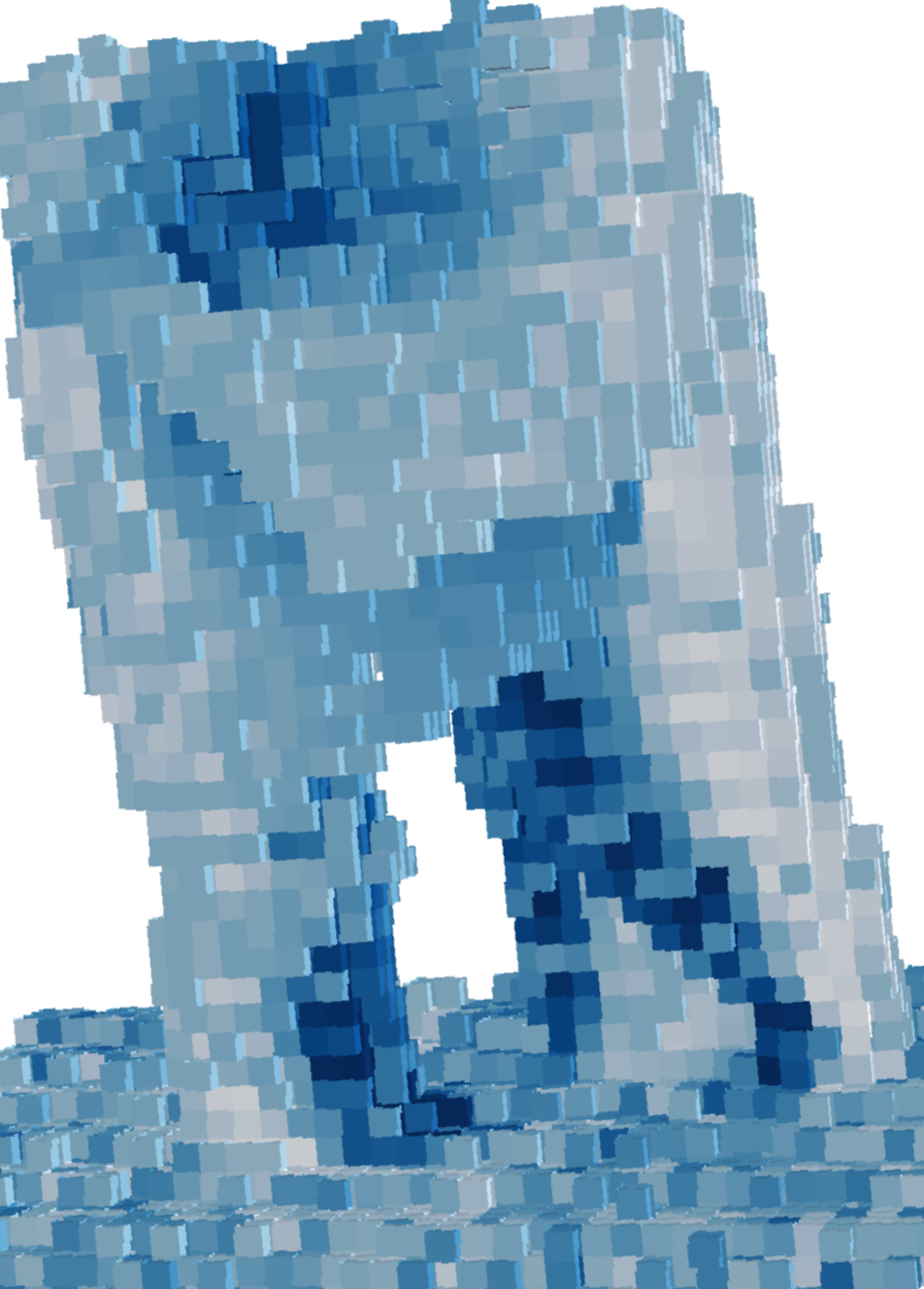}}
\subfigure[Pushing] {\label{fig:di_example(c)}
\includegraphics[height=2.4cm]{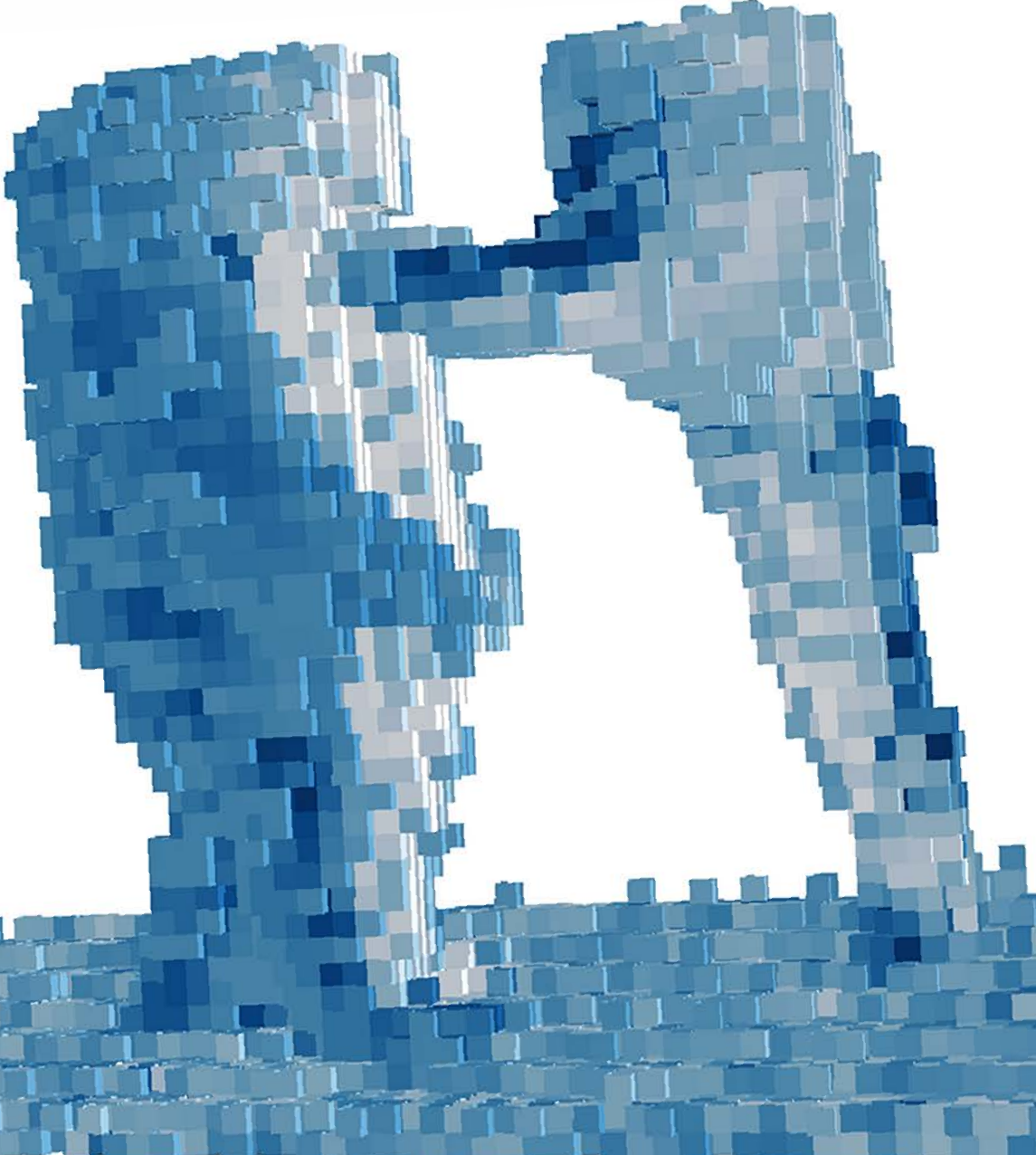}}
\caption{The 3DV examples from NTU RGB+D 60 dataset~\cite{shahroudy2016ntu}.}
\label{fig:3dv_example}
\vspace{-0.2cm}
\end{figure}
\subsection{3DV extraction using temporal rank pooling}
With the binary 3D appearance voxel sets above, temporal rank pooling is executed to generate 3DV. A linear temporal ranking score function will be defined for compressing the voxel sets into one voxel set (i.e., 3DV).

Particularly, suppose $V_i ,\dots, V_T$ indicate the binary 3D appearance voxel sets, and $\overline{V_i} = \frac{1}{t}\times\sum_{i}^{t}V_i$ is their average till time $t$. The  ranking score function at time $t$ is given by
\begin{equation}
\label{eq:ranking_score}
S(t|\textbf{w})=\left\langle \textbf{w}, \overline{V_i}\right\rangle,
\end{equation}
where $\textbf{w}\in\mathbb{R}^d$ is the ranking parameter vector. $\textbf{w}$ is learned from the depth video to reflect the ranking relationship among the frames. The criteria is that, the later frames are of larger ranking scores as
\begin{equation}
\label{eq:ranking_relationship}
q>t \Rightarrow S(q|\textbf{w})>S(t|\textbf{w}).
\end{equation}
The learning procedure of $\textbf{w}$ is formulated as a convex optimization problem using RankSVM~\cite{smola2004tutorial} as
\begin{equation}
\begin{aligned}
\textbf{w}^*&=\mathop{\text{argmin}}\limits_\textbf{w} \frac{\lambda}{2}\parallel \textbf{w} \parallel^2+\\ &{\frac{2}{T(T-2)}\times \sum_{q>t} \text{max}\left\{0,1-S(q|\textbf{w})+S(t|\textbf{w})\right\}}.
\end{aligned}
\label{eq:rankSVM}
\end{equation}
Specifically, the first term is the often used regularizer for SVM. And, the second is the hinge-loss for soft-counting how many pairs $q>t$ are incorrectly ranked, which does not obey $S(q|\textbf{w})>S(t|\textbf{w})+1$. Optimizing Eqn.~\ref{eq:rankSVM} can map the 3D appearance voxel sets  $V_i,\cdot\cdot\cdot,V_T$ to a single vector $\textbf{w}^*$. Actually, $\textbf{w}^*$ encodes the dynamic evolution information from all the frames. Spatially reordering $\textbf{w}^*$ from 1D to 3D in voxel form can construct 3DV for 3D action characterization. Thus, each 3DV voxel can be jointly encoded by the corresponding $\textbf{w}^*$ item as motion feature and its regular 3D position index $\left(x,y,z\right)$ as spatial feature. Some more 3DV examples are shown in Fig.~\ref{fig:3dv_example}. We can intuitively observe that, 3DV can actually distinguish the different actions from motion perspective even human-object or human-human interaction happens. Meanwhile to accelerate 3DV extraction for application, the approximated temporal rank pooling~\cite{bilen2017action} is used by us during implementation.

\begin{figure}
\centering
\includegraphics[width=0.4\textwidth]{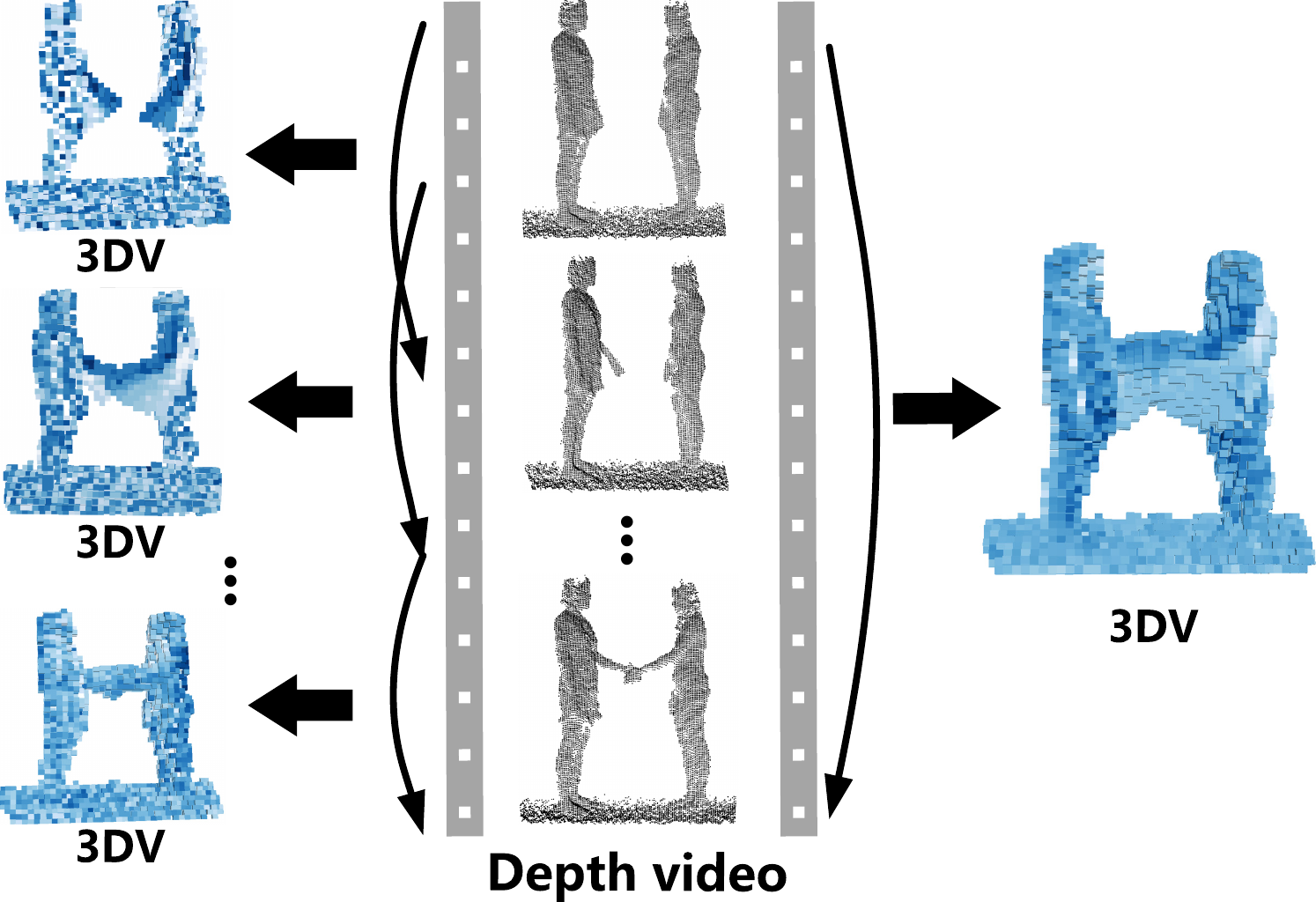}
\caption{Temporal split for 3DV extraction.}
\label{fig:TemporalSegment}
\vspace{-0.2cm}
\end{figure}
\subsection{Temporal split} \label{sec:temporal_segmentation}
Applying temporal rank pooling to whole depth video may vanish some fine temporal order information. To better maintain motion details, we propose to execute temporal split for 3DV. The depth video will be divided into $T_1$ temporal splits with the overlap ratio of 0.5, which is the same as~\cite{xiao2019action}. 3DV will extract from all the temporal splits and the whole depth video simultaneously as in Fig.~\ref{fig:TemporalSegment}, to involve the global and partial temporal 3D motion clues jointly.
\subsection{Action proposal}
Since background is generally not helpful for 3D action characterization, action proposal is also conducted by us, following~\cite{xiao2019action} but with some minor modifications. First YOLOv3-Tiny~\cite{redmon2018yolov3} is used for human detection instead of Faster R-CNN~\cite{ren2015faster}, concerning running speed. Meanwhile, human and background are separated by depth thresholding. Particularly, depth value histogram is first extracted with the discretization interval of 100 mm. The interval of highest occurrence probability is then found. The threshold is empirically set as its mediate value plus 200 mm. Then, 3DV will be extracted only from action proposal's 3D space.
\section{Deep learning network on 3DV}
After acquiring 3DV, the upcoming problem is how to conduct deep learning on it to conduct feature learning and 3D action type decision jointly. Since 3DV appears in 3D voxel form, an intuitive way is to apply 3D CNN to it as many 3D visual recognition methods~\cite{maturana2015voxnet,ge20173d,moon2018v2v} does. Nevertheless 3D CNN is generally hard to train, mainly due to its relatively large number of model parameters. Deep learning on point set (e.g., PointNet~\cite{qi2017pointnet} and PointNet++~\cite{qi2017pointnet++}) is the recently emerged research avenue to address the disordered characteristics of point set, with promising performance and lightweight model size. Inspired by this, we propose to apply PointNet++ to conduct deep learning on 3DV instead of 3D CNN concerning effectiveness and efficiency jointly. To this end, 3DV will be abstracted into point set form. To our knowledge, using PointNet++ to deal with voxel data has not been well studied before. Meanwhile since 3DV tends to loose some appearance information as shown Fig.~\ref{fig:pointcloud_voxel}, a multi-stream deep learning model based on PointNet++ is also proposed to learn appearance and motion feature for 3D action characterization.

\begin{figure}
\centering
\includegraphics[width=0.41\textwidth]{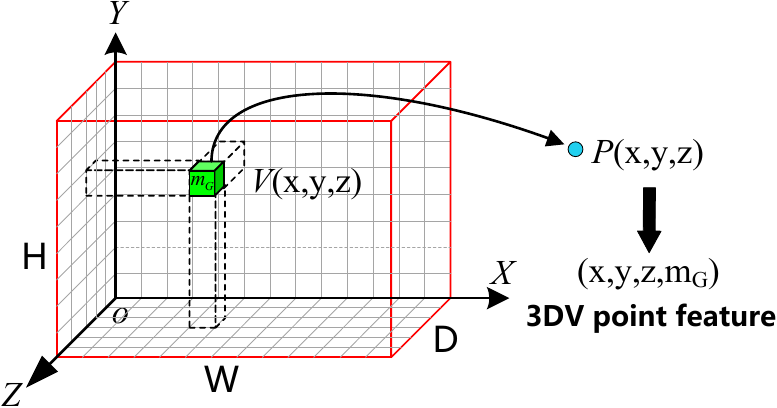}
\caption{The procedure of abstracting 3DV voxel $V\left\{x,y,z\right\}$ into 3DV point $P\left\{x,y,z\right\}$.}
\label{fig:voxel2point}
\vspace{-0.2cm}
\end{figure}

\subsection{Review on PointNet++}\label{PC_Classification}
PointNet++~\cite{qi2017pointnet++} is derived from PointNet~\cite{qi2017pointnet}, the pioneer in deep learning on point set. PointNet is proposed mainly to address the disordered problem within point clouds. However, it cannot capture the local fine-grained pattern well. PointNet++ alleviates this in a local-to-global hierarchical learning manner. It declares 2 main contributions. First, it proposes to partition the set of points into overlap local regions to better maintain local fine 3D visual clue. Secondly, it uses PointNet recursively as the local feature learner. And, the local features will be further grouped into larger units to reveal the global shape characteristics. In summary, PointNet++ generally inherits the merits of PointNet but with stronger local fine-grained pattern descriptive power. {Compared with 3D CNN, PointNet++ is generally of more light-weight model size and higher running speed. Meanwhile, it tends to be easier to train.}


The main intuitions for why we apply PointNet++ to 3DV lie into 3 folders. First, we do not want to trap in the training challenges of 3D CNN. Secondly, PointNet++ is good at capturing local 3D visual pattern, which is beneficial for 3D action recognition. That is, local 3D motion pattern actually plays vita role for good 3D action characterization, as the hand region shown in Fig.~\ref{fig:3dv} towards ``Handshaking". Last, applying PointNet++ to 3DV is not a difficult task. What we need to do is to abstract 3DV into the point set form, which will be illustrated next.


\subsection{Abstract 3DV into point set}\label{PC_abstract}

Suppose the acquired 3DV for a depth video without temporal split is of size $H\times W \times D$, each 3DV voxel $V\left(x,y,z\right)$ will possesses a global motion value $m_G$ given by temporal rank pooling as in Fig.~\ref{fig:voxel2point} where $\left(x,y,z\right)$ indicates the 3D position index of $V\left(x,y,z\right)$ within 3DV. To fit PointNet++, $V\left(x,y,z\right)$ is then abstracted as a 3DV point $P\left(x,y,z\right)$ with the descriptive feature of $\left(x,y,z,m_G\right)$. Particularly, $\left(x,y,z\right)$ denotes the 3D spatial feature and $m$ is the motion feature. Thus, the yielded $P\left(x,y,z\right)$ is able to represent the 3D motion pattern and corresponding spatial information integrally. Since $\left(x,y,z\right)$ and $m$ are multi-modular feature, feature normalization is executed to balance their effect towards PointNet++ training. Specifically, $m$ is linearly normalized into the range of $\left[-0.5,0.5\right]$. Towards the spatial feature, $y$ is first linearly normalized into the range of $\left[-0.5,0.5\right]$. Then $x$ and $z$ are re-scaled respectively, according to their size ratio towards $y$. In this way, the 3D geometric characteristics can be well maintained to alleviate distortion. As illustrated in Sec.~\ref{sec:temporal_segmentation}, temporal split is executed in 3DV to involve multi-temporal motion information. Thus, each 3DV point $P\left(x,y,z\right)$ will correspond to multiple global and local motion values. We propose to concatenate all the motion values to describe 3D motion pattern integrally. $P\left(x,y,z\right)$ will be finally characterized by the spatial-motion feature as
\begin{equation}
F_{P\left(x,y,z\right)}=(\overbrace{x,y,z}^{Spatial}
                                            \overbrace{m_G,m_1,...m_{T_1}}^{Motion}) \,,
\end{equation}
where $m_G$ is the motion feature extracted from whole video; $m_i$ denotes motion feature from the $i$-th temporal split; and $T_1$ is the number of temporal splits as in Sec.~\ref{sec:temporal_segmentation}.

\begin{figure}
  \centering
  \includegraphics[width=0.48\textwidth]{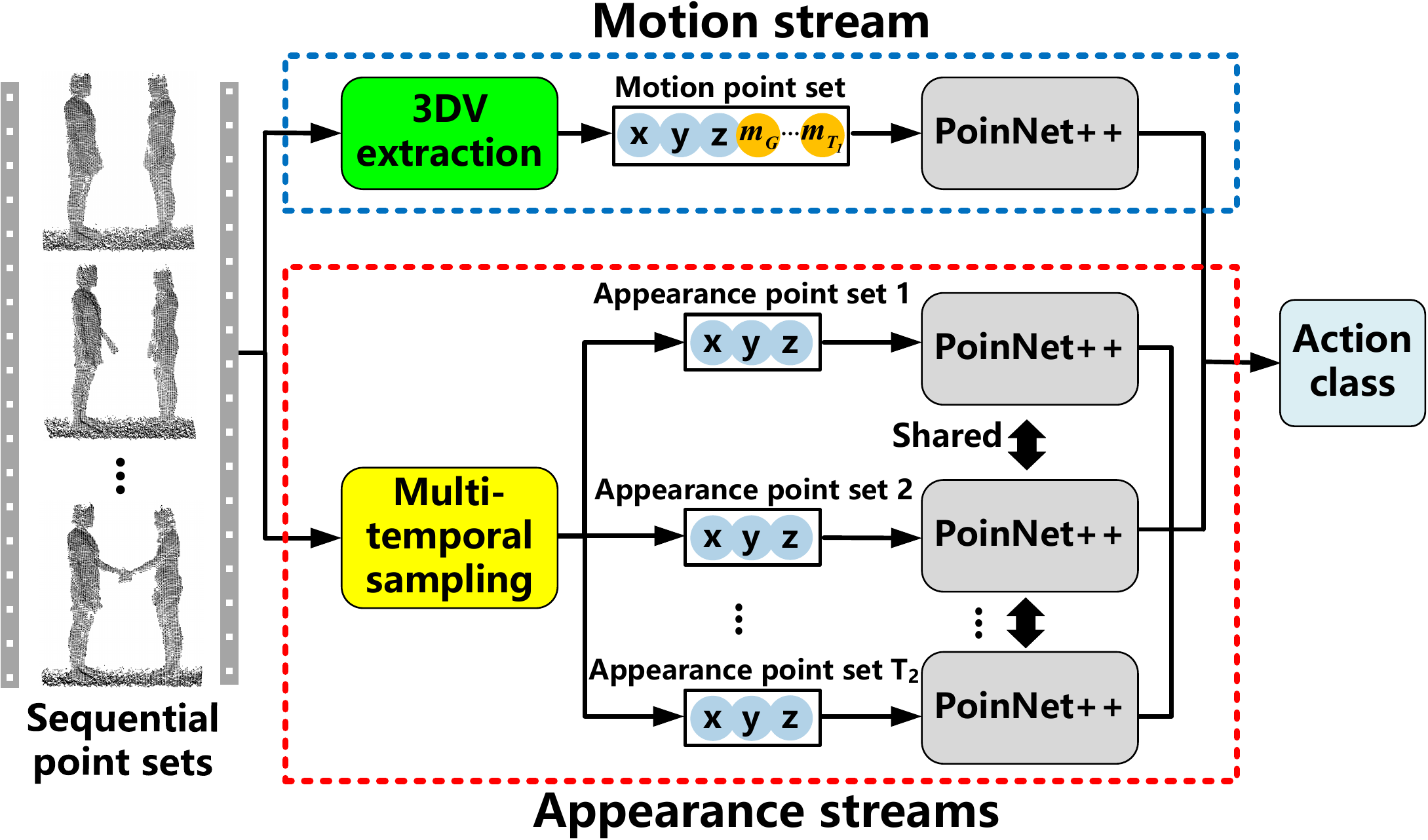}
  \caption{PointNet++ based multi-stream network for 3DV to learn motion and appearance feature jointly.}\label{fig:framework}
  \vspace{-0.2cm}
\end{figure}
\subsection{Multi-stream network}\label{multistreams}
Since 3DV may lose fine appearance clue, a multi-stream network using PointNet++ is proposed to learn motion and appearance feature jointly, following the idea in~\cite{simonyan2014two} for RGB video. As in Fig.~\ref{fig:framework}, it consists of 1 motion stream and multiple appearance streams. The input of motion stream is the single 3DV point set from Sec.~\ref{PC_abstract}. For motion PointNet++ the 3DV points with all the motion features of 0 will not be sampled. And, the inputs of appearance streams are the raw depth point sets sampled from $T_2$ temporal splits with action proposal. Particularly, they share the same appearance PointNet++. Motion and appearance feature is late fused via concatenation at fully-connected layer.
\section{Implementation details}
3DV voxel is set of size $35mm \times 35mm\times 35mm$. $T_1$ and $T_2$ is set to 4 and 3 respectively, for multi-temporal motion and appearance feature extraction. For PointNet++, farthest point sampling is used on the centroids of local regions. The sampled points are grouped with ball query. The group radius at the first and second level is set to 0.1 and 0.2 respectively. Adam~\cite{kingma2014adam} is applied as the optimizer with batch size of 32. Leaning rate begins with 0.001, and decays with a rate of 0.5 every 10 epochs. Training will end when reaching 70 epochs. During training, we perform data augmentation for 3DV points and raw depth points including random rotation around $Y$ and $X$ axis, jittering and random points dropout. Multi-stream network is implemented using PyTorch. Within each stream, PointNet++ will sample 2048 points for both of motion and appearance feature learning.



\section{Experiments}
\subsection{Experimental setting}

{\bf Dataset: NTU RGB+D 120}~\cite{liu2019ntu}. It is the most recently emerged challenging 3D action recognition dataset, and also of the largest size. Particularly, 114,480 RGB-D action samples of 120 categories captured using Microsoft Kinect v2 are involved in this dataset. These involved action samples are of large variation on subject, imaging viewpoint and background. This imposes essential challenges to 3D action recognition. The accuracy of the state-of-the-art approaches is not satisfactory {(i.e., below $70\%$)} both under the cross-subject and cross-setup evaluation criteria.

{\bf Dataset: NTU RGB+D 60}~\cite{shahroudy2016ntu}. It is the preliminary version of NTU RGB+D 120. That is, 56,880 RGB-D action samples of 60 categories captured using Microsoft Kinect v2 are involved in this dataset. Before NTU RGB+D 120, it is the largest 3D action recognition dataset. Cross-subject and cross-view evaluation criteria is used for test.

{\bf Dataset: N-UCLA}~\cite{wang2014cross}. Compared with NTU RGB+D 120 and NTU RGB+D 120, this is a relatively small-scale 3D action recognition dataset. It only contains 1475 action samples of 10 action categories. These samples are captured using Microsoft Kinect v1 from 3 different viewpoints, with relatively higher imaging noise. Cross-view evaluation criteria is used for test.

{\bf Dataset: UWA3DII}~\cite{rahmani2016histogram}. This is also a small-scale 3D action recognition dataset with only 1075 video samples from 30 categories. One essential challenge of this dataset is the limited number of training samples per action category. And, the samples are captured using Microsoft Kinect v1 with relatively high imaging noise.

{\bf Input data modality and evaluation metric}. During experiments, the input data of our proposed 3DV based 3D action recognition method is only depth maps. We will not use any other auxiliary information, such as skeleton, RGB image, human mask, etc. The training / test sample splits and testing setups on all the 4 datasets are strictly followed for fair comparison. Classification accuracy on all the action samples is reported for performance evaluation.

\begin{table}[t]
	\caption{Performance comparison on action recognition accuracy ($\%$) among different methods on NTU RGB+D 120 dataset.}
\vspace{0.05cm}
	\centering
	\scriptsize
	\begin{tabular}{l|c|c}
		\Xhline{0.8pt}
		Methods                              & Cross-subject & Cross-setup    \\ \hline
        \multicolumn{3}{c}{\textbf{Input: 3D Skeleton}}                    \\ \hline
        NTU RGB+D 120 baseline~\cite{liu2019ntu}              & 55.7             &57.9         \\ \hline
        GCA-LSTM~\cite{liu2017global}       & 58.3            &59.3 \\\hline
        FSNet~\cite{liu2019skeleton}                         & 59.9          &62.4 \\ \hline
        Two stream attention LSTM~\cite{liu2017skeleton} & 61.2            &63.3 \\\hline
        Body Pose Evolution Map ~\cite{liu2018recognizing} & 64.6            &66.9 \\\hline
        SkeleMotion~\cite{caetano2019skelemotion}&67.7  &66.9               \\\hline
        \multicolumn{3}{c}{\textbf{Input: Depth maps}}                      \\ \hline
		NTU RGB+D 120 baseline~\cite{liu2019ntu}              & 48.7             &40.1          \\ \hline
        3DV-PointNet++ (ours)               &{\bf 82.4}           &{\bf93.5}                \\ \Xhline{0.8pt}
	\end{tabular}
	\label{table:NTU120result}
\end{table}

\begin{table}[t]
	\caption{Performance comparison on action recognition accuracy ($\%$) among different methods on NTU RGB+D 60 dataset.}
\vspace{0.05cm}
	\centering
	\scriptsize
	\begin{tabular}{l|c|c}
		\Xhline{0.8pt}
		Methods                              & Cross-subject & Cross-view    \\ \hline

        \multicolumn{3}{c}{\textbf{Input: 3D Skeleton}}                    \\ \hline
        SkeleMotion~\cite{caetano2019skelemotion}&69.6  &80.1 \\\hline
        GCA-LSTM~\cite{liu2017global}   &74.4 &82.8 \\\hline
        Two stream attention LSTM~\cite{liu2017skeleton} & 77.1            &85.1 \\\hline

        AGC-LSTM~\cite{si2019attention} &89.2& 95.0            \\\hline
        AS-GCN~\cite{li2019actional}    &86.8&94.2  \\\hline
        VA-fusion~\cite{zhang2019view}  &89.4 &95.0\\\hline
        2s-AGCN~\cite{shi2019two}       &88.5&95.1  \\\hline
        DGNN~\cite{shi2019skeleton}     &{\bf 89.9}&96.1  \\\hline

        \multicolumn{3}{c}{\textbf{Input: Depth maps}}                      \\ \hline
	    HON4D~\cite{oreifej2013hon4d}	      &30.6	     &7.3          \\ \hline
		SNV~\cite{yang2014super}              &31.8      &13.6           \\ \hline
        HOG$^2$	~\cite{ohn2013joint}	      &32.2		 &22.3 		     \\ \hline
		Li.~\cite{li2018unsupervised}         & 68.1      &83.4          \\ \hline
		Wang.~\cite{Wang2018Depth}            & 87.1             &84.2          \\ \hline
		MVDI~\cite{xiao2019action}            & 84.6             &87.3          \\ \hline
        3DV-PointNet++ (ours)                &88.8     &{\bf 96.3}                  \\\Xhline{0.8pt}
	\end{tabular}
	\label{table:NTU60result}
\end{table}

\subsection{Comparison with state-of-the-art methods}

{\bf NTU RGB+D 120:} Our 3DV based approach is compared with the state-of-the-art skeleton-based and depth-based 3D action recognition methods~\cite{liu2019ntu,liu2017global,liu2017skeleton,liu2018recognizing,caetano2019skelemotion} on this dataset. The performance comparison is listed in Table~\ref{table:NTU120result}. We can observe observed that:

$\bullet$ It is indeed impressive that, our proposition achieves the breaking-through results on this large-scale challenging dataset both towards the cross-subject and cross-setup test settings. Particularly we achieve $82.4\%$ and $93.5\%$ on these 2 settings respectively, which outperforms the state-of-the-art manners by large margins (i.e., $14.7\%$ at least on cross-subject, and $26.6\%$ at least on cross-setup). This essentially verifies the superiority of our proposition;

$\bullet$ The performance of the other methods is poor. This reveals the great challenges of NTU RGB+D 120 dataset;


$\bullet$ Our method achieves better performance on cross-setup case than cross-subject. This implies that, 3DV is more sensitive to subject variation.

\begin{table}[t]
	\caption{Performance comparison on action recognition accuracy ($\%$) among different depth-based methods on N-UCLA dataset.}
\vspace{0.05cm}
	\centering
	\scriptsize
	\begin{tabular}{l|c}
		\Xhline{0.8pt}
		Methods                              & Accuracy    \\ \hline
        HON4D~\cite{oreifej2013hon4d}       &39.9           \\ \hline
        SNV~\cite{yang2014super}            & 42.8          \\ \hline
        AOG~\cite{wang2014cross}               & 53.6       \\ \hline
        HOPC~\cite{rahmani2014hopc}           &80.0           \\ \hline
        MVDI~\cite{xiao2019action}            & 84.2           \\ \hline
        3DV-PointNet++ (ours)               &{\bf 95.3} \\ \Xhline{0.8pt}
	\end{tabular}
	\label{table:N-UCLAresult}
\end{table}

{\bf NTU RGB+D 60:} The proposed method is compared with the state-of-the-art approaches ~\cite{liu2017global,liu2017skeleton,si2019attention,li2019actional,zhang2019view,shi2019two,shi2019skeleton,oreifej2013hon4d,yang2014super,ohn2013joint,li2018unsupervised,Wang2018Depth,xiao2019action} on this dataset. The performance comparison is listed in Table~\ref{table:NTU60result}. We can see that:

$\bullet$ Our proposition still significantly outperforms all the depth-based manners, both on the cross-subject and cross-view test settings.


$\bullet$ On cross-view setting, the proposed method is also superior to all the skeleton-based manners. And, it is only slightly inferior to DGNN~\cite{shi2019skeleton} on cross-subject setting; This reveals that, only using depth maps can still achieve the promising performance.

$\bullet$ By comparing Table~\ref{table:NTU120result} and~\ref{table:NTU60result}, we can find that the performance of some methods (i.e., GCA-LSTM~\cite{liu2017global}, Two stream attention LSTM~\cite{liu2017skeleton},) significantly drops. Concerning the shared cross-subject setting, GCA-LSTM drops $16.1\%$ and Two stream attention LSTM drops $15.9\%$. However, our manner only drops $6.4\%$. This demonstrates 3DV's strong adaptability and robustness.

\begin{table}[t]
	\caption{Performance comparison on action recognition accuracy ($\%$) among different depth-based methods on UWA3DII dataset.}
\vspace{0.05cm}
	\centering
	\scriptsize
	\begin{tabular}{l|c}
		\Xhline{0.8pt}
		Methods                              &Mean accuracy    \\ \hline
        HON4D~\cite{oreifej2013hon4d}       &28.9           \\ \hline
        SNV~\cite{yang2014super}            & 29.9          \\ \hline
        AOG~\cite{wang2014cross}               & 26.7       \\ \hline
        HOPC~\cite{rahmani2014hopc}           &52.2           \\ \hline
        MVDI~\cite{xiao2019action}            & 68.1           \\ \hline
        3DV-PointNet++ (ours)                &{\bf 73.2}        \\ \Xhline{0.8pt}
	\end{tabular}
	\label{table:UWA3Dresult}
\end{table}

{\bf N-UCLA and UWA3DII:} We compared the proposed manner with the state-of-the-art depth-based approaches~\cite{oreifej2013hon4d,yang2014super,wang2014cross,rahmani2014hopc,xiao2019action} on these 2 small-scale datasets. The performance comparison is given in Table~\ref{table:N-UCLAresult} and~\ref{table:UWA3Dresult} respectively. To save space, the average accuracy of the different viewpoint combinations is reported on UWA3DII. It can be summarized that:

$\bullet$ On these 2 small-scale datasets, the proposed approach still consistently outperforms the other depth-based manners. This demonstrates that, our proposition takes advantages over both of the large-scale and small-scale test cases;

$\bullet$ 3DV does not perform well on UWA3DII, with the accuracy of $73.2\%$. In our opinion, this may be caused by the fact that the training sample amount per class is limited on this dataset. Thus, deep learning cannot be well conducted.

\subsection{Ablation study}

\begin{table}[t]
	\caption{Effectiveness of 3DV motion feature on NTU RGB+D 120 dataset. Appearance stream is not used.}
\vspace{0.05cm}
	\centering
	\scriptsize
	\begin{tabular}{c|c|c|c}
		\Xhline{0.8pt}
         $T_1$    & 3DV point feature    & Cross-subject & Cross-setup \\ \hline
        1    & ($x,y,z$)            &    61.4       &  68.9  \\
		1    &$\left(x,y,z,m_G\right)$        &    75.1       &  87.4     \\\Xhline{0.8pt}
	\end{tabular}
	\label{table:MotionFeature}
\end{table}

{\bf Effectiveness of 3DV motion feature:} To verify this, we choose to remove 3DV motion feature from the sampled 3DV points within PointNet++ to observe the performance change. The comparison results on NTU RGB+D 120 dataset are given in Table~\ref{table:MotionFeature}. We can see that, without the motion feature 3DV's performance will significantly drop (i.e., $ 18.5\%$ at most).


\begin{table}[t]
	\caption{Effectiveness of temporal split for 3DV extraction on NTU RGB+D 120 dataset. Appearance stream is not used.}
\vspace{0.05cm}
	\centering
	\scriptsize
	\begin{tabular}{c|c|c|c}
		\Xhline{0.8pt}
         $T_1$    & 3DV point feature    & Cross-subject & Cross-setup \\ \hline
		1    &$\left(x,y,z,m_G\right)$        &    75.1       &  87.4     \\
        2    &$\left(x,y,z,m_G,m_1,m_2\right)$  &    75.8       &  89.6  \\
        4    &$\left(x,y,z,m_G,m_1,\dots,m_4\right)$  &    76.9       &  92.5      \\ \Xhline{0.8pt}
	\end{tabular}
	\label{table:TemporalMotion}
\end{table}

{\bf Effectiveness of temporal split for 3DV extraction:} Towards this, the temporal split number $T_1$ is set to 1, 2, and 4 respectively on NTU RGB+D 120 dataset. The comparison results are listed in Table~\ref{table:TemporalMotion}. Obviously, temporal split can essentially leverage the performance in all test cases.

\begin{table}[t]
\centering
\scriptsize
\caption{Effectiveness of appearance stream on NTU RGB+D 60 and 120 dataset.}
\vspace{0.05cm}
\label{table:appearance}

\begin{tabular}{c|c|c|c}
\Xhline{0.8pt}
Dataset                                               & Input stream      & Cross-subject  & Cross-setup \\ \hline
\multicolumn{1}{l|}{\multirow{3}{*}{\text{NTU 120}}}  & 3DV                & 76.9       & 92.5       \\ \cline{2-4}
\multicolumn{1}{c|}{}                                 & Appearance       & 72.1          & 79.4   \\ \cline{2-4}
\multicolumn{1}{l|}{}                                 & 3DV+appearance & {\bf82.4}       & {\bf 93.5}       \\ \hline
                                               &       & Cross-subject  & Cross-view \\ \hline
\multicolumn{1}{c|}{\multirow{3}{*}{\text{NTU 60}}}   & 3DV                & 84.5       & 95.4      \\ \cline{2-4}
\multicolumn{1}{c|}{}                                 & Appearance       & 80.1          & 85.1    \\ \cline{2-4}
\multicolumn{1}{c|}{}                                 & 3DV+appearance & {\bf 88.8}       & {\bf 96.3}        \\ \Xhline{0.8pt}
\end{tabular}%
\end{table}

{\bf Effectiveness of appearance stream:} This is verified on NTU RGB+D 60 and 120 dataset simultaneously, as listed in Table~\ref{table:appearance}. We can observe that:

$\bullet$ The introduction of appearance stream can consistently enhance the performance of 3DV on these 2 datasets, towards all the 3 test settings;

$\bullet$ 3DV stream significantly outperforms the appearance stream consistently, especially on cross-setup and cross-view settings. This verifies 3DV's strong discriminative power for 3D action characterization in motion way.

\begin{table}[t]
	\caption{Effectiveness of action proposal on N-UCLA dataset.}
\vspace{0.05cm}
	\centering
	\scriptsize

	\begin{tabular}{cc}
		\Xhline{0.8pt}
        Action proposal   & Accuracy  \\ \hline
         W/O&  92.9   \\ \hline
           With& {\bf95.3}                \\ \Xhline{0.8pt}
	\end{tabular}
	\label{table:actionproposal}
\end{table}

{\bf Effectiveness of Action proposal:} The performance comparison of our method with and without action proposal on N-UCLA dataset is listed in Table~\ref{table:actionproposal}. Actually, action proposal can help to enhance performance.

{\bf PointNet++ vs. 3D CNN:} To verify the superiority of PointNet++ for deep learning on 3DV, we compare it with 3D CNN. particularly, the well-established 3D CNN model (i.e., C3D~\cite{tran2015learning}) for video classification is used with some modification. That is, the number of C3D's input channels is reduced from 3 to 1. And, 3DV is extracted with the fixed size of $50\times50\times50$ grids as the input of C3D. Without data augmentation and temporal split, the performance and model complexity comparison on N-UCLA and NTU RGB+D 60 dataset (cross-view setting) is given in Table~\ref{table:3DCNN}. We can see that, PointNet++ essentially takes advantage both on effectiveness and efficiency.

\begin{table}[t]
	\caption{Comparison on performance and complexity between PointNet++ and C3D on N-UCLA and NTU RGB+D 60 dataset.}
\vspace{0.05cm}
	\centering
	\scriptsize
	\begin{tabular}{ccccc}
		\Xhline{0.8pt}
        Method         &Parameters   &FLOPs         & N-UCLA & NTU RGB+D 60\\ \hline
         C3D           &29.2M        &10.99G        &64.5     &85.0 \\ \hline
         PointNet++    &{\bf1.24M}   &{\bf1.24G}    &{\bf71.3}&{\bf90.0}     \\ \Xhline{0.8pt}
	\end{tabular}
	\label{table:3DCNN}
\end{table}


\subsection{Parameter analysis}

\begin{table}[t]
	\caption{Performance comparison among the different sampling point numbers for 3DV, on NTU RGB+D 120 dataset.}
\vspace{0.05cm}
	\centering
	\scriptsize
	\begin{tabular}{c|c|c}
		\Xhline{0.8pt}
		Sampling point number    &   Cross-subject    &   Cross-view  \\ \hline
		512    &74.9           &    89.0      \\ \hline
        1024   &75.7 & 90.9    \\ \hline
        2048   &{\bf76.9} &{\bf92.5}          \\ \Xhline{0.8pt}
	\end{tabular}
	\label{table:pointsnumbers}
\end{table}

\begin{table}[t]
	\caption{Performance comparison among the different 3DV voxel sizes, on NTU RGB+D 120 dataset.}
\vspace{0.05cm}
	\centering
	\scriptsize
	\begin{tabular}{c|c|c}
		\Xhline{0.8pt}
		Voxel size (mm) &Cross-subject &Cross-view    \\ \hline
        $25\times25\times25$   &75.9      &92.0                      \\ \hline
        $35\times35\times35$   &{\bf76.9}      &{\bf92.5}                   \\ \hline
        $50\times50\times50$   &76.0      &91.6                   \\ \hline
        $75\times75\times75$   &74.1      &90.4                   \\\Xhline{0.8pt}
	\end{tabular}
	\label{table:voxelsize}
\end{table}

{\bf Sampling point number on 3DV:} Before inputting 3DV point set into PointNet++, farthest point sampling is executed first. To investigate the choice of sampling point number, we compare the performance of 3DV stream with the different sampling point number values on NTU RGB+D 120 dataset. The results are listed in Table~\ref{table:pointsnumbers}. That is, 2048 can achieve the best performance on 3DV.

{\bf 3DV voxel size:} To investigate the choice of 3D voxel size, we compare the performance of 3DV stream with the different 3DV voxel sizes on NTU RGB+D 120 dataset. The results are listed in Table~\ref{table:voxelsize}. Particularly, $35mm\times35mm\times35mm$ is the optimal 3DV voxel size.

\subsection{Other issues}
{\bf Running time:} On the platform with CPU: Intel(R) Xeon(R) CPU E5-2690 v3 @ 2.6GHz (only using 1 core), and GPU: 1 Nvidia RTX 2080Ti, 3DV's overall online running time is 2.768s/video as detailed in Table~\ref{tab:time}. Particularly, 100 samples with the average length of 97.6 frames are randomly selected from NTU RGB+D 60 dataset for test.

{\bf Approximated temporal rank pooling:} In our implementation, the approximated temporal rank pooling is used for 3DV extraction due to its high running efficiency. We compare it with the original one on N-UCLA data, using CPU. As shown in Table~\ref{table:rankpool}, the approximated temporal rank pooling runs much faster than the original one with the similar performance.

\begin{table}[t]
\scriptsize
\caption{Item-wise time consumption of 3DV per video.}
\centering
\begin{tabular}{@{\hspace{0.5mm}}c@{\hspace{0.5mm}}|@{\hspace{0.5mm}}c@{\hspace{0.5mm}}|@{\hspace{0.5mm}}c@{\hspace{0.5mm}}
||@{\hspace{0.5mm}}c@{\hspace{0.5mm}}|@{\hspace{0.5mm}}c@{\hspace{0.5mm}}|@{\hspace{0.5mm}}c@{\hspace{0.5mm}}}
\Xhline{0.8pt} Unit &  Item & Time (ms) & Unit & Item & Time (ms)\\ \hline
GPU & Human detection  & 231ms & CPU & 3DV Pointlization & 88ms \\ \hline
CPU & Point cloud voxelization & 2107ms  & GPU & PointNet++ forward  & 242ms\\ \hline
CPU & Temporal rank pooling  & 100ms  &\multicolumn{2}{c|@{\hspace{0.5mm}}}{\textbf{Overall}} & 2768ms\\ \Xhline{0.8pt}
\end{tabular}
\label{tab:time}
\end{table}

\begin{table}[t]
	\caption{Performance comparison between the original and approximated temporal rank pooling for 3DV on N-UCLA dataset.}
    \vspace{0.05cm}
	\centering
	\scriptsize
	\begin{tabular}{ccc}
		\Xhline{0.8pt}
        Temporal rank pooling methods   & Accuracy & Time per sample \\ \hline
        Original   & {\bf95.8}   &    1.12s       \\ \hline
        Approximated   &95.3   &   {\bf0.10s}           \\ \Xhline{0.8pt}
	\end{tabular}
	\label{table:rankpool}
\end{table}

\begin{figure}
  \centering
  \includegraphics[width=0.48\textwidth]{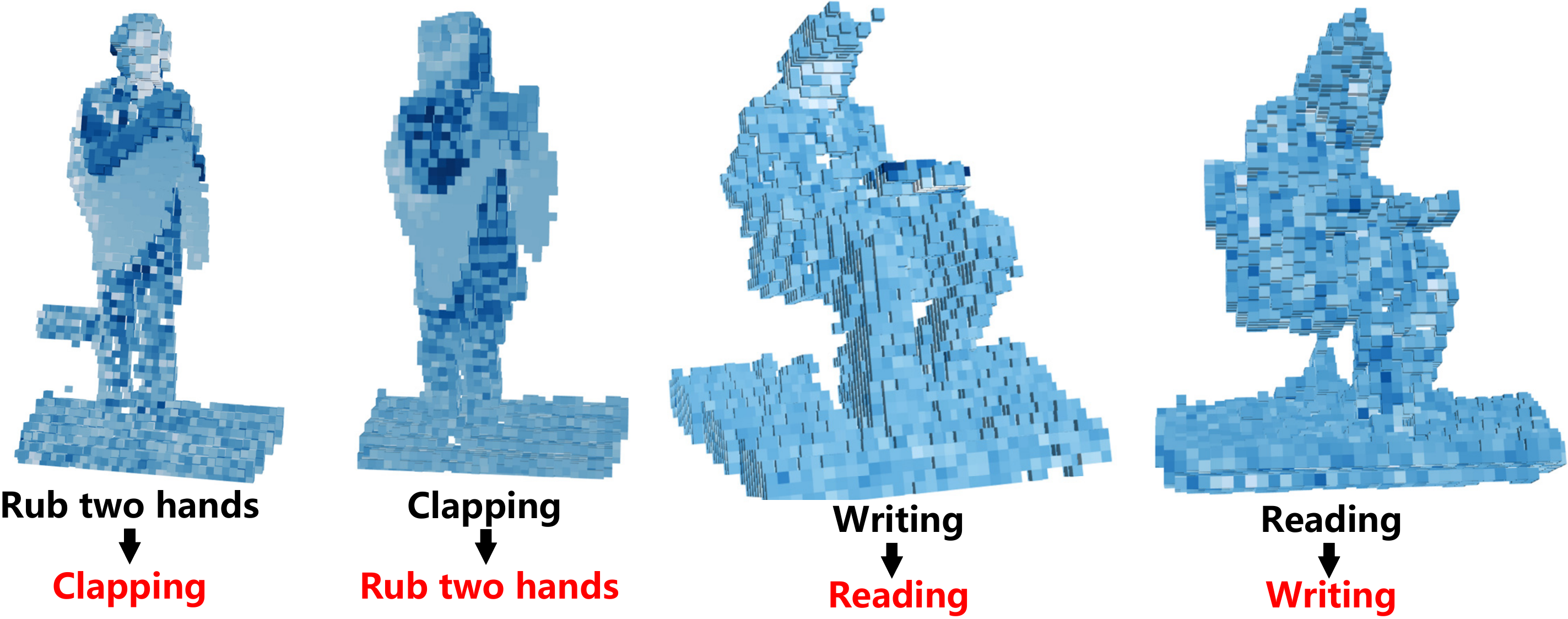}
  \caption{Some classification failure cases of 3DV. Ground-truth action label is shown in black, and the prediction is in red.}\label{fig:failure_cases}
  \vspace{-0.2cm}
\end{figure}

{\bf 3DV failure cases:} Some classification failure cases of 3DV are shown in Fig.~\ref{fig:failure_cases}. We find that, the failures tend to be caused by the tiny motion difference between the actions.

\section{Conclusions}
In this paper, 3DV is proposed as a novel and compact 3D motion representation for 3D action recognition. PointNet++ is applied to 3DV to conduct end-to-end feature learning. Accordingly, a multi-stream PointNet++ based network is also proposed to learn the 3D motion and depth appearance feature jointly to better characterize 3D actions. The experiments on 4 challenging datasets demonstrate the superiority of our proposition both for the large-scale and small-scale test cases. How to further enhance 3DV's discriminative power is what we mainly concern about in future, especially towards the tiny motion patterns.

\section*{Acknowledgment}
This work is jointly supported by the National Natural Science Foundation of China (Grant No. 61502187 and 61876211),  Equipment Pre-research Field Fund of China (Grant No. 61403120405), the Fundamental Research Funds for the Central Universities (Grant No. 2019kfyXKJC024), National Key Laboratory Open Fund of China (Grant No. 6142113180211),  the start-up funds from University at Buffalo. Joey Tianyi Zhou is supported by Singapore Government's Research, Innovation and Enterprise 2020 Plan (Advanced Manufacturing and Engineering domain) under Grant A1687b0033 and Grant A18A1b0045.

{\small
\bibliographystyle{ieee_fullname}
\bibliography{egbib}
}

\end{document}